\documentclass[letterpaper]{article} % DO NOT CHANGE THIS
\usepackage{aaai2026}  % DO NOT CHANGE THIS
\usepackage{times}  % DO NOT CHANGE THIS
\usepackage{helvet}  % DO NOT CHANGE THIS
\usepackage{courier}  % DO NOT CHANGE THIS
\usepackage[hyphens]{url}  % DO NOT CHANGE THIS
\usepackage{graphicx} % DO NOT CHANGE THIS
\usepackage{natbib}  % DO NOT CHANGE THIS AND DO NOT ADD ANY OPTIONS TO IT
\usepackage{caption} % DO NOT CHANGE THIS AND DO NOT ADD ANY OPTIONS TO IT
\usepackage{algorithm}
\usepackage{algorithmic}

\usepackage{pifont}
\usepackage{multirow}
\usepackage{bbding}
\usepackage{xcolor}
\usepackage{colortbl, booktabs}
\usepackage{amssymb}
\usepackage{amsmath}
\usepackage{newfloat}
\usepackage{listings}
\DeclareCaptionStyle{ruled}{labelfont=normalfont,labelsep=colon,strut=off} % DO NOT CHANGE THIS
\floatstyle{ruled}
\newfloat{listing}{tb}{lst}{}
\floatname{listing}{Listing}
\title{Invisible Triggers, Visible Threats! Road-Style Adversarial Creation Attack for Visual 3D Detection in Autonomous Driving}
\author{
    Jian Wang, Lijun He, Yixing Yong, Haixia Bi, Fan Li\textsuperscript{\thanks{Corresponding author.}}\\
}
\affiliations{
    School of Information and Communications Engineering, Xi’an Jiaotong University\\
    wj851329121@stu.xjtu.edu.cn, lijunhe@mail.xjtu.edu.cn, yongyx@stu.xjtu.edu.cn, haixia.bi@mail.xjtu.edu.cn, lifan@mail.xjtu.edu.cn

}

\usepackage{bibentry}
\begin{document}

\maketitle

\begin{abstract}
Modern autonomous driving (AD) systems leverage 3D object detection to perceive foreground objects in 3D environments for subsequent prediction and planning. Visual 3D detection based on RGB cameras provides a cost-effective solution compared to the LiDAR paradigm. While achieving promising detection accuracy, current deep neural network-based models remain highly susceptible to adversarial examples. The underlying safety concerns motivate us to investigate realistic adversarial attacks in AD scenarios.
Previous work has demonstrated the feasibility of placing adversarial posters on the road surface to induce hallucinations in the detector. However, the \emph{unnatural appearance} of the posters makes them easily noticeable by humans, and their \emph{fixed content} can be readily targeted and defended. 
To address these limitations, we propose the AdvRoad to generate diverse road-style adversarial posters. The adversaries have naturalistic appearances resembling the road surface while compromising the detector to perceive non-existent objects at the attack locations. We employ a two-stage approach, termed Road-Style Adversary Generation and Scenario-Associated Adaptation, to maximize the attack effectiveness on the input scene while ensuring the natural appearance of the poster, allowing the attack to be carried out stealthily without drawing human attention. 
Extensive experiments show that AdvRoad generalizes well to different detectors, scenes, and spoofing locations. Moreover, physical attacks further demonstrate the practical threats in real-world environments.
\end{abstract}

\begin{links}
    \link{Code}{https://github.com/WangJian981002/AdvRoad}
    %\link{Datasets}{https://aaai.org/example/datasets}
    %\link{Extended version}{https://aaai.org/example/extended-version}
\end{links}

\section{Introduction}
\begin{figure}[!t]
	\centering
	\includegraphics[width=\linewidth]{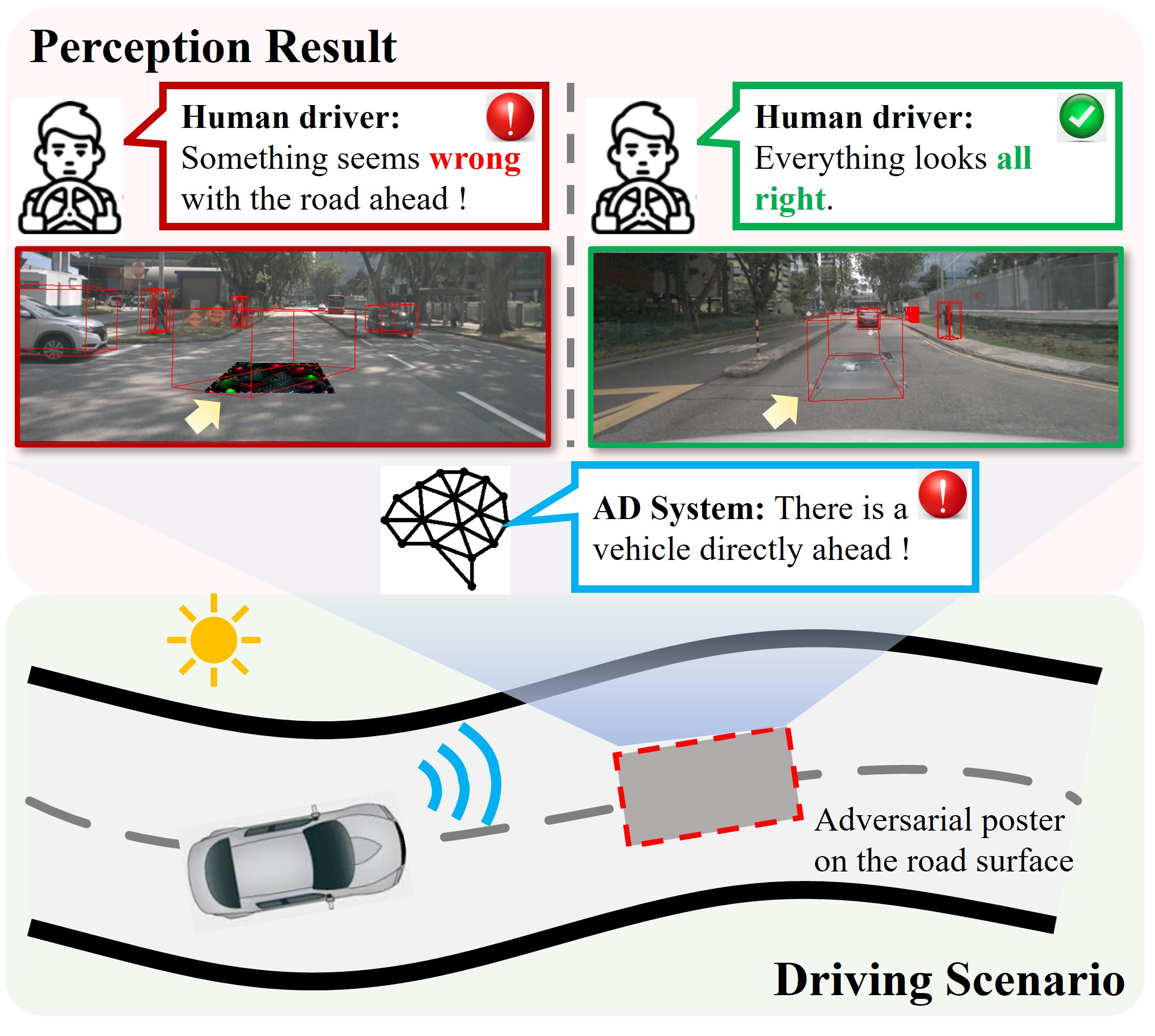}
	\caption{Illustration of the adversarial FP attacks on the road. The 3D detection system will perceive a ghost object near the poster. Compared with previous work (left), our poster (right) is harder to attract human attention, making it more likely to pose a real threat.}
	\label{fig_fisrt_page}
	%\Description{A woman and a girl in white dresses sit in an open car.}
\end{figure}

Visual 3D object detection \cite{ma2024vision,hu2023planning,chen2017multi} has emerged as a pivotal technology in autonomous driving systems \cite{mao20233d,chen2024end,gulino2023waymax}, offering cost-effective environmental perception through widely accessible RGB cameras. Despite its computational efficiency and hardware affordability compared to LiDAR-dependent methods, the reliability of deep neural networks (DNNs) in safety-critical scenarios remains questionable due to their vulnerability to adversarial attacks \cite{yang2025adversarial,lin2024hard,han2023interpreting,wang2023towards}. 
Recent studies have revealed that carefully designed inputs, such as additive perturbations \cite{zhang2024visual,zhang2021evaluating,goodfellow2014explaining,athalye2018synthesizing} and local patches \cite{guesmi2024dap,thys2019fooling,hu2024adversarial,hu2025two}, can catastrophically alter the output of DNNs, and these adversarial examples can be successfully implemented in physical scenarios \cite{sato2021dirty}. This security threat poses unpredictable consequences given the life-critical nature of 3D perception systems, which motivates us to investigate realistic adversarial attacks for 3D object detection in real-world environments.  

\begin{table*}[h]
	\centering
	\begin{tabular}{c|cc}
		\toprule
		\textbf{Victim Model}&\textbf{Attack Way}&\textbf{Consequence}\\ 
		\midrule
		\multirow{5}*{LiDAR-based}&Malicious laser signal injection &\multirow{2}*{FN}\\
		&\cite{cao2023you,jin2023pla,hau2021object}&\\\cline{2-3}
		&Malicious laser signal injection&\multirow{2}*{FP}\\
		&\cite{jin2023pla,sun2020towards,cao2019adversarial,wang2023adversarial}\\\cline{2-3}
		&3D adversarial mesh \cite{tu2020physically,cao2019adversarialv2}&FN\\
		\midrule
		\multirow{4}*{Camera-based}&Adversarial camouflage \cite{li2023adv3d}&FN\\\cline{2-3}
		&Adversarial patch&\multirow{2}*{FN}\\
		&\cite{zhu2023understanding,wang2025unified,cheng2023fusion,xie2023adversarial}\\\cline{2-3}
		&Adversarial poster \cite{wang2025physically}&FP\\
		%&Side 
		%&\multicolumn{4}{c}{\textbf{Scene Creating Attack}}
		
		\bottomrule
	\end{tabular}
	
	\caption{The summarization of physical adversarial attacks targeting 3D object detection in AD.}
	\label{tab:related_works}
\end{table*}

Adversarial attacks on 3D object detectors can be divided into two types based on model errors: false negatives (FN) where real objects evade detection, and false positives (FP) where non-existent targets are identified. Most existing studies \cite{zhu2023understanding,abdelfattah2021adversarial,cheng2023fusion,tu2020physically,zhang2024comprehensive,wang2025unified} focus on FN attacks --- for example, attaching adversarial patches to vehicles to make them invisible to detectors, which may result in a rear-end collision. However, implementing FN attacks typically requires physical access to the target object, limiting their practical application. FP attacks, on the other hand, aim to make detectors "see" imaginary obstacles, potentially triggering sudden braking or dangerous lane changes by autonomous vehicles. Despite posing similar safety risks as FN attacks, FP attacks for visual 3D detection remain poorly studied in the current attack literature.

Recently, Wang \emph{et al.} \cite{wang2025physically} pioneered physical FP attack targeting visual 3D detectors by placing a carefully optimized poster on the road, thereby inducing the detector to perceive a ghost object near the poster (as shown in Fig. \ref{fig_fisrt_page} left). Since the poster is 2D and lacks thickness, it is flexible to print and launch the attack. Moreover, the generated poster has strong generalization ability, allowing it to be effective in various scenarios.
They learn the poster by proposing an image-3D applying algorithm that can differentiably render the 3D space poster onto the image, and directly optimizing the poster's pixel values. \textbf{However}, the following weaknesses remain: \ding{182} \emph{The content of the poster significantly differs from the road surface, making it easily noticed by humans.} Directly optimizing poster pixels cannot constrain the learned content, which often has patterns with non-naturalistic appearances. \ding{183} \emph{Each training session can only generate one poster, making it easy to be targeted and defended.} They generate a single adversarial poster that is effective across all scene images. Despite achieving a high attack success rate, the single poster can be easily exploited and defended against (e.g., by fine-tuning the model with data containing this adversary).

In response to limitations \ding{182} and \ding{183}, we propose a novel FP attack pipeline that generates diverse road-style adversarial posters. All posters have patterns similar to the background road surface, making them harder to attract human attention, allowing the attack to proceed silently. Our framework employs a two-stage approach, Road-Style Adversary Generation and Scenario-Associated Adaptation, to generate diverse naturalistic adversaries. In the first
stage, we utilize GAN-based techniques to train an adversarial generator that maps latent noise vectors to road-style adversaries. First, we use drones to take aerial photos of ground scenes and construct a road image collection for training the style discriminator. Then, we gradually update the generator by iteratively rendering the mapped poster onto the scene image and backpropagating the gradients based on the adversarial objective and style discriminator. The first stage ensures that the generated posters are similar to the source collection images while being able to compromise the detector. In the second stage, we derive a  local-optimal poster tailored to a specific input, aiming to enhance its deceptive effectiveness within the given scenario. Concretely, we initialize a poster from the generator trained in the first stage, randomly sample and place it at various locations in the current scene, and then backpropagate the gradients to the latent space to optimize the noise vector toward a locally optimal solution.
Note that in this stage, the generator is frozen. Therefore, the found adversary not only has naturalistic road styles but also achieves the strongest attack effectiveness in the current scene.

In short, our contributions are:
\begin{itemize}
	\item We present a naturalistic FP attack pipeline for inducing the ghost object on the road. The crafted posters can significantly evade human perception while compromising the 3D models, thereby increasing the practical threat to the AD system.
	\item We introduce Road-Style Adversary Generation and Scenario-Associated Adaptation to maximize the attack capability of the adversarial poster. Moreover, all adversaries are effective across various scenes and at certain observation distances (e.g., $\leq 10m$).
	\item Extensive experiments in both the digital and physical worlds demonstrate the effectiveness of our approach with improved stealthiness. In addition, our posters are harder to defend against using existing defense techniques.
\end{itemize}

\begin{figure*}[!t]
	\centering
	\includegraphics[width=5.1in]{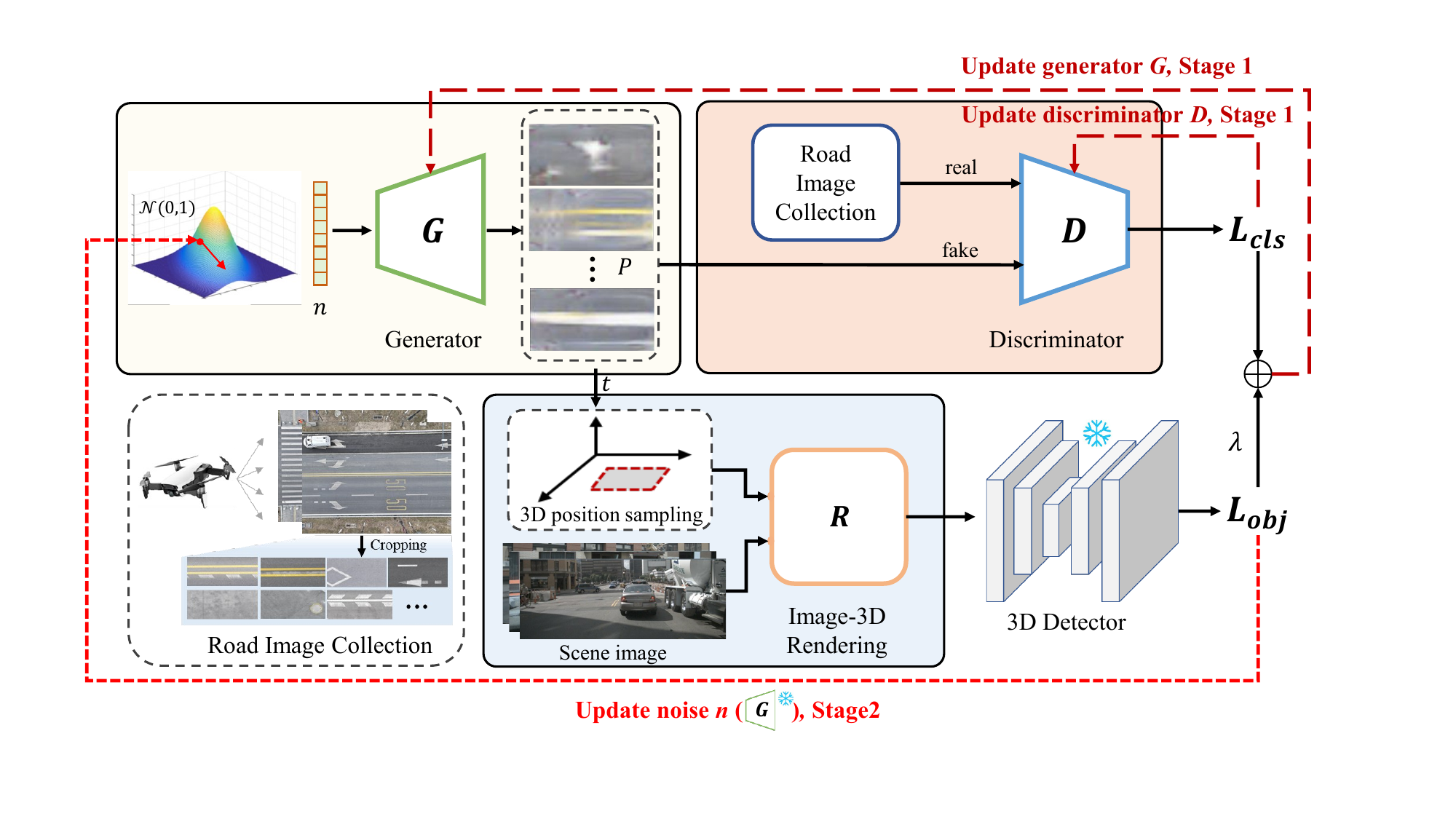}
	\caption{The AdvRoad framework. Stage 1 trains an adversarial generator that outputs universal road-style posters; Stage 2 updates the poster (the latent vector) to enhance the attack capability for the given scene.}
	\label{fig_overall}
	%\Description{A woman and a girl in white dresses sit in an open car.}
\end{figure*}

\section{Related Work}

3D object detection is the most crucial perception task in modular autonomous driving systems, which identifies and locates surrounding traffic participants like vehicles and pedestrians in 3D space. Errors in detection results gradually accumulate, thereby affecting subsequent predictions and planning.

Recent research efforts have developed various physical attack methods targeting LiDAR-based and camera-based detection algorithms. We provide a summary of these works in terms of physical attack ways and attack consequences in Table \ref{tab:related_works}. 
For LiDAR-based approaches, attackers must alter the captured LiDAR point clouds. This can be achieved by strategically emitting laser pulses toward the target LiDAR sensor \cite{cao2019adversarial,jin2023pla,cao2023you} or placing 3D adversarial objects in the environment \cite{tu2020physically,cao2019adversarialv2}. However, altering LiDAR data typically requires complex equipment like laser diodes, photodiodes, or industrial-grade 3D printers. This limits the attack's flexibility in dynamically changing AD scenarios. For camera-based approaches, attackers can utilize more affordable adversarial patches to perturb the captured images, leading to various detection errors. 
Extending Wang \emph{et al.}'s framework \cite{wang2025physically} of using road surface posters to create fake 3D objects, we further study how to hide these patterns from human drivers. The spoofing capability to detectors and stealthiness to humans make the attack more dangerous in real driving scenarios.

\section{Problem Definition}
We investigate adversarial FP attacks against visual 3D object detection models by placing the learned poster on the road surface. Specifically, attackers apply the adversarial poster to a benign image $x$ by firstly sampling the poster location in 3D space and then rendering the poster onto the image. This results in an adversarial input $\mathcal{R}(x,\delta,t)$, where $\mathcal{R}(\cdot)$ is the rendering function that applies the adversary $\delta$ to the input $x$ according to the sampled transformation parameter $t$.
Our goal is to induce the 3D detector, denoted as $F_{\theta}$, to produce a desired response $y^*$ at the poster position, which is formally defined by maximizing the likelihood of the response $y^*$ under the adversarial input:
\begin{equation} 
	%\label{xxx}
	\mathop{\max}_{\delta} \log p(y^*|\mathcal{R}(x,\delta,t))
\end{equation}
where $p$ is the probability function; $\delta$ is the adversary to be optimized, which can either be an explicit representation of the poster (e.g., a pixel array $P \in [0,255]^{3\times H\times W}$) or an implicit representation (e.g., a generator $G: n\in \mathbb{R}^d\rightarrow P \in [0,255]^{3\times H\times W}, n\sim \mathcal{N}(0, 1)$). 

\textbf{Continuous Attack Goals.} When driving, the observation of the scene changes continuously, such as varying distances and viewing angles. The practical threat posed by a road-surface poster is substantially reduced if its effectiveness is confined to a specific scenario or location. Because the AD system is likely to classify targets appearing in merely one or two frames as sensor noise. Consequently, these posters must exhibit universality to remain effective across varying scenarios and viewing distances.

To this end, we leverage the Expectation of Transformation (EoT) \cite{athalye2018synthesizing} across all training samples and a wide range of spoofing locations to enhance the physical robustness and generality of the attack. Formally, the optimization objective becomes:
\begin{equation}
	\delta = \arg\max \sum_{x\in X}^{}\sum_{t\in T}^{}\log p(y^*|\mathcal{R}(x,\delta,t))
\end{equation}
where $X$ comprises all possible image inputs and $T$ is the set of position transformation parameters.

\section{Proposed AdvRoad Framework}
\subsection{Road-Style Adversary Generation}
Fig. \ref{fig_overall} presents the framework of our road-style creation attack.
The first stage aims to learn an adversarial generator that outputs diverse road-style posters while containing critical 
foreground features, enabling them to induce FP predictions in the detector. We employ a standard GAN pipeline to integrate road surface style and spoofing information to the generator.  

\textbf{Road Image Collection.} We utilize a DJI drone to capture aerial photography of traffic scenes, obtaining a series of raw images containing diverse road surfaces. We then crop authentic road patches from the collected aerial images, with each patch dimension approximating vehicle size (2m$\times$4m). The built collection comprises over 2,000 road surface images covering various road patterns and styles (as shown in Fig. \ref{fig_overall} left). This collection serves as the real reference for training the style discriminator, while the synthetic counterparts are generated by the generator. The inclusion of authentic imagery facilitates learning the realistic road patterns from the generator, thereby enhancing the visual indistinguishability of adversarial outputs to human drivers.

\textbf{Adversarial Objective.} The adversarial generator $G$ is trained on the whole detection dataset $X$. Given a frame of input image $x\in X$, we first sample the poster locations in the 3D space then render the poster $G(n), n\in \mathbb{R}^d$ to the $x$ through differentiable Image-3D rendering (discussed later). Then, we prepare the adversarial label $y^*$ for the input $\mathcal{R}(x,\delta,t)$. We mask out the regions of real objects on the input image using ground truth (GT) bounding box annotations, and replace the GT labels with spoofing bounding boxes introduced by the posters. This approach provides dual advantages: First, masking surrounding objects prevents the gradient being vanishing by averaging. Second, we can reuse the detector's native loss function to directly calculate the adversarial loss. Formally, the adversarial loss is:
\begin{equation}
	L_{obj} = J(F_{\theta}(\mathcal{R}(x,G(n),t)), y^*)
\end{equation}
where $J(\cdot)$ is the original loss function for detector $F_{\theta}$. The other training objective comes from style discriminator $D$. We freeze the $D$ and maximize the discriminator's confidence in classifying the generator's outputs as real. The final training objective for the generator is:
\begin{equation}
	L_{G} = L_{cls}+ \lambda \cdot L_{obj}
\end{equation}
We alternately train the generator and the discriminator, gradually injecting spoofing and style information into the generated posters.

\subsection{Scenario-Associated Adaptation}
After the first stage training, the posters output by the generator both resemble the road surface and maintain a certain degree of deception. They can serve as a universal trap to trigger FP predictions in the 3D detector. However, the generated posters rely on sampling noise from the latent space---a process that involves inherent randomness and may lead to unstable attack effects. Therefore, we further introduce Scenario-Associated Adaptation to enhance the attack capability of the posters in specific scenarios.

Specifically, given the input scenario $x$, we first \emph{randomly initialize} the noise vector $n$ from the Gaussian distribution $\mathcal{N}(0, 1)$ and fed $n$ into the \emph{frozen} generator to get the spoofing poster.  
Then, we randomly sample the poster locations in the current scene and perform rendering. Finally, we back-propagate the gradient to the latent space based on the adversarial loss $L_{obj}$ and update vector $n$. This process is formulated as:
\begin{equation}
	\begin{aligned}
		&n_0 = n\\
		&n_{i+1} = n_i -\alpha\cdot \nabla_{n_i}J(F_{\theta}(\mathcal{R}(x,G(n_i),t)), y^*)
	\end{aligned}
\end{equation}
We repeat this process until reach the maximum iteration number. To preserve realism, we ensure that the updated noise $n_{i+1}$ falls into the hypersphere of radius $\eta$ centered at initial noise $n_0$ in each iteration.
The final poster exhibits strongest deceptive capability in the current scene while being authentic.

\subsection{Image-3D Rendering}
\label{section_Image_3D_Rendering}
Conventional 2D patch rendering \cite{thys2019fooling,lee2019physical,hu2021naturalistic} solely performs scaling and rotating the patch according to the objects' 2D bounding boxes. Although convenient, the physical size of the patch and its position in 3D space are not considered, which are critical for realistic physical attacks on 3D detectors. Following \cite{wang2025physically,zhu2023understanding}, we briefly introduce how to render the poster from the 3D road surface onto the image. 

First, we sample the 3D spatial positions of posters to place them on the road surface. Within a sector spanning $\pm \Delta_{\theta}$ relative to the vehicle's heading direction and a distance range of $d_{min}$ to $d_{max}$ meters, we randomly sample placement locations while avoiding overlapping with existing scene objects. Second, we project the four poster corner points onto the image plane using the camera's intrinsic and extrinsic matrices, thereby determining the adversarial region in the image (the quadrangle region defined by projected corner points). Third, for each pixel within this region, we inversely calculate its 3D coordinates aided by road height information (approximated from the bottom face height of the nearest scene object to the poster). Finally, each pixel's RGB value is calculated via bilinear interpolation based on its 3D position relative to the poster. For more details refer to supplementary-A.

\section{Experiment}
\subsection{Experimental Setup}
\textbf{Victim Model.} The vision-based BEV space 3D detector BEVDet \cite{huang2021bevdet}, BEVDet4D \cite{huang2022bevdet4d}, and BEVFormer \cite{li2024bevformer} are selected as victim models for the attack, considering their representativeness and the fact that BEV space inherently supports most downstream perception tasks for AD \cite{hu2023planning}.
%BEVDet and BEVDet4D both leverage geometry-based transformation to project the perspective view (PV) image features to the BEV, while the latter further integrates historical frames' information. The BEVFormer leverages transformer architecture to aggregate valuable image features for pre-defined BEV grids. 
For each detector, the ResNet50 \cite{he2016deep} and SwinTransformer-Tiny \cite{liu2021swin} are used as image backbone respectively.

\textbf{Dataset.} For the digital attack, we use nuScenes dataset \cite{caesar2020nuscenes} to train the adversary and perform the attack. nuScenes is a large-scale, multi-modal dataset specifically designed for AD and 3D object detection. The training and validation set contains 28,130 and 6,019 frames respectively, with each frame including image data from six cameras and $360^{\circ}$ 3D object annotations. We train all detectors on the training set following their official settings and detection performances are given in supplementary-B.1. The confidence scores for the detected objects are over 0.1 for all models following \cite{wang2025physically,tu2020physically}.

\textbf{Evaluation Metric.} The attack success rate (ASR) is used to evaluate the creation attack, which measures the proportion of successfully detected fake objects among all spoofing attempts. We consider a fake object is successfully detected when the minimum distance between the detector's predictions and the center of the poster is less than $d_{thr}$. Multiple center distance thresholds, $\{2.0m, 1.5m, 1.0m, 0.5m\}$, are adopted for comprehensive evaluation. Unlike using the IoU as an indicator, precise alignment of the detected bounding box with the $y^*$ in terms of size and orientation is less critical. Since the AD system may lead to dangerous consequences as long as a fake obstacle is perceived near the poster.

Moreover, we use the Learned Perceptual Image Patch Similarity (LPIPS) score \cite{zhang2018unreasonable} to assess the environmental consistency of the attack, which measures perceptual similarity between benign and attacked images. 

\textbf{Implementation Details.}
We set the category of the spoofing object to the most common \emph{vehicle}, with the poster's physical size being $2m\times 4m$.  
For spoofing locations, we aim to induce FP predictions either in front of or behind the self-vehicle. Therefore, the posters are placed at a distance of 7 to 10 meters from the self-vehicle with $\Delta_{\theta}=5^{\circ}$. Within this range, the AD system's misjudgment leaves little time for the driver to react. We sample 1,000 validation frames and place posters at two locations per frame, yielding 2,000 attacks for ASR computation. See the supplementary materials for more training details.

\subsection{Main Result}

\begin{figure}[!t]
	\centering
	\includegraphics[width=\linewidth]{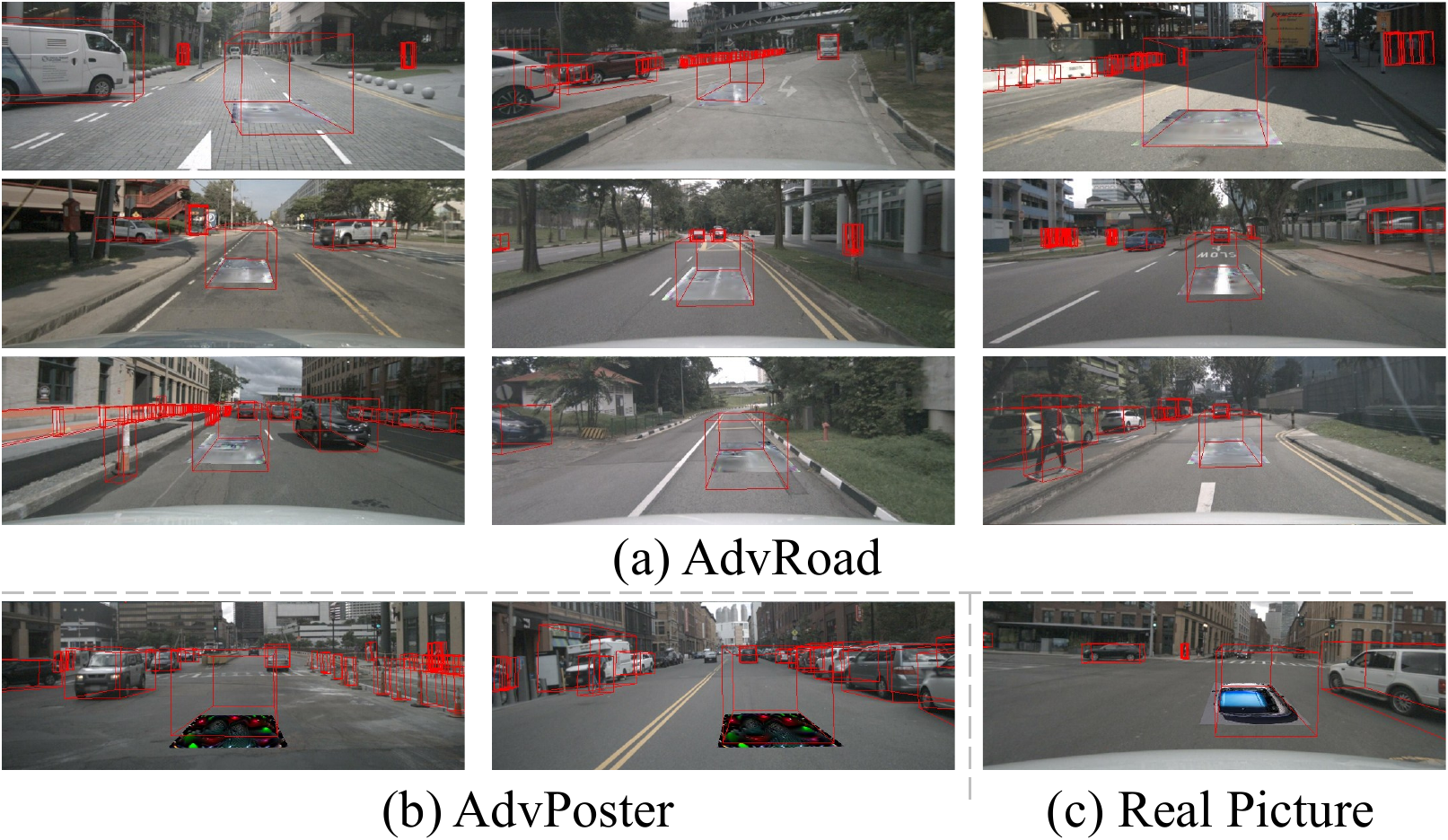}
	\caption{Visualizations of attack results in the digital domain. We place the spoofing poster on the road surface to launch the attack. (1) AdvRoad, our road-style naturalistic adversarial poster; (2) AdvPoster \cite{wang2025physically}, generated by directly optimizing the pixel values; (3) Real Picture, use images of real vehicles as posters.}
	\label{fig_main}
	%\Description{A woman and a girl in white dresses sit in an open car.}
\end{figure}

\begin{table}[!t]
	\centering
	\scriptsize
	\begin{tabular}{l|c|c|p{0.45cm}<{\centering}p{0.45cm}<{\centering}p{0.45cm}<{\centering}p{0.45cm}<{\centering}}
		\toprule
		\multirow{2}*{\textbf{Model}}&\multirow{2}*{\textbf{Attack}}&\multirow{2}*{\textbf{LPIPS $\downarrow$}}&\multicolumn{4}{c}{\textbf{ASR ($\%, \uparrow$)}}\\ 
		&&&$\mathbf{2.0m}$&$\mathbf{1.5m}$&$\mathbf{1.0m}$&$\mathbf{0.5m}$\\
		\midrule
		&Benign&-&$1.5$&$0.3$&$0.1$&$0$\\
		BEVDet&Random&$0.2136$&$8.0$&$4.9$&$2.9$&$0.8$\\
		-R50&Real picture&$0.2066$&$30.4$&$22.8$&$15.7$&$6.3$\\
		&\cellcolor{gray!20} \textbf{AdvRoad}&\cellcolor{gray!20} $0.1472$&\cellcolor{gray!20} $62.6$&\cellcolor{gray!20} $55.6$&\cellcolor{gray!20} $42.7$&\cellcolor{gray!20} $23.3$\\
		\midrule
		&Benign&-&$1.2$&$0.2$&$0$&$0$\\
		BEVDet&Random&$0.2138$&$1.7$&$0.7$&$0.4$&$0.1$\\
		-SwinT&Real picture&$0.2066$&$25.7$&$19.0$&$12.1$&$4.9$\\
		&\cellcolor{gray!20} \textbf{AdvRoad}&\cellcolor{gray!20} $0.1337$&\cellcolor{gray!20} $60.2$&\cellcolor{gray!20} $56.3$&\cellcolor{gray!20} $47.6$&\cellcolor{gray!20} $28.8$\\
		\midrule
		&Benign&-&$1.2$&$0.1$&$0$&$0$\\
		BEVDet4D&Random&$0.2137$&$3.4$&$1.9$&$1.2$&$0.3$\\
		-R50&Real picture&$0.2066$&$45.1$&$38.1$&$29.0$&$14.9$\\
		&\cellcolor{gray!20} \textbf{AdvRoad}&\cellcolor{gray!20} $0.1331$&\cellcolor{gray!20} $49.1$&\cellcolor{gray!20} $42.9$&\cellcolor{gray!20} $32.7$&\cellcolor{gray!20} $17.7$\\
		\midrule
		&Benign&-&$1.2$&$0.3$&$0$&$0$\\
		BEVDet4D&Random&$0.2136$&$1.4$&$0.3$&$0$&$0$\\
		-SwinT&Real picture&$0.2066$&$23.4$&$19.3$&$13.7$&$7.3$\\
		&\cellcolor{gray!20} \textbf{AdvRoad}&\cellcolor{gray!20} $0.1370$&\cellcolor{gray!20} $39.1$&\cellcolor{gray!20} $35.1$&\cellcolor{gray!20} $27.8$&\cellcolor{gray!20} $15.7$\\
		\midrule
		&Benign&-&$1.4$&$0.3$&$0$&$0$\\
		BEVFormer&Random&$0.1304$&$1.4$&$0.3$&$0$&$0$\\
		-R50&Real picture&$0.1260$&$6.3$&$3.5$&$1.8$&$0.7$\\
		&\cellcolor{gray!20} \textbf{AdvRoad}&\cellcolor{gray!20}$0.0822$&\cellcolor{gray!20}$44.5$&\cellcolor{gray!20}$32.7$&\cellcolor{gray!20}$20.7$&\cellcolor{gray!20}$8.2$\\
		\midrule
		&Benign&-&$1.5$&$0.3$&$0$&$0$\\
		BEVFormer&Random&$0.1306$&$1.5$&$0.5$&$0.1$&$0$\\
		-SwinT&Real picture&$0.1260$&$20.9$&$16.6$&$10.4$&$4.3$\\
		&\cellcolor{gray!20} \textbf{AdvRoad}&\cellcolor{gray!20}$0.0818$&\cellcolor{gray!20}$37.3$&\cellcolor{gray!20}$30.6$&\cellcolor{gray!20}$21.0$&\cellcolor{gray!20}$8.9$\\
		\bottomrule
		%&Side 
		%&\multicolumn{4}{c}{\textbf{Scene Creating Attack}}

	\end{tabular}
	\caption{Digital attack results of adversarial creation attack in the nuScenes dataset.}
	\label{tab:main_results}
\end{table}

We verify the effectiveness of our road-style adversarial poster (AdvRoad) in comparison with Benign, Random, and Real picture. Specifically, \emph{Benign:} original scene without the adversarial pattern, however, we still sample and record the attack locations (with a fixed random seed) considering the miscalculation cases, where the detection results for the real objects are incorrectly counted to spoofed ones. \emph{Random:} randomly initialize a poster for the attack. \emph{Real picture:} use images of real vehicles as posters for the attack (Fig. \ref{fig_main}(c)).

Table \ref{tab:main_results} shows the digital attack results in nuScenes dataset. We achieve good attack performance across six different 3D detectors, successfully inducing FP predictions near the poster locations (see Fig. \ref{fig_main}(a) for the visualization results). This holds regardless of whether the target model uses a CNN-based or Transformer-based backbone, a geometry-based or network-based PV-to-BEV transformation, or an anchor point-based or query-based detection head. For AD systems, such an error rate (exceeding $~40\%$) is catastrophic. A suddenly appearing object within 10 meters in front of the ego-vehicle can trigger emergency braking or lane-changing decisions, potentially leading to severe safety incidents. Moreover, since our posters resemble the road surface, drivers have limited time to react and effectively intervene.

Since we avoid overlapping with scene objects when sampling attack locations, miscalculation cases are nearly negligible, e.g., the ASR for Benign consistently $<1.5\%$ under $CD_{2.0m}$. Moreover, we observe an interesting result that taking a real vehicle image as the spoofing poster may also lead to FP errors. Specifically, Real picture can achieve an ASR of up to $45.1\%$ under $CD_{2.0m}$ for BEVDet4D-R50.
Although the picture poster is 2D and lacks thickness, it can still provide intrinsic foreground visual cues that the detector captures and recognizes. However, similar to a real object, a `flat' vehicle is also likely to attract the driver's attention and has higher LPIPS scores compared with AdvRoad.

\subsection{Comparison with Current Work}
We compare our attack with AdvPoster \cite{wang2025physically} in terms of attack performance, naturality, and defense difficulty, which explicitly learns the adversary by directly optimizing the pixel values of the poster.

\begin{figure}[t]
	\centering
	\includegraphics[width=0.95\linewidth]{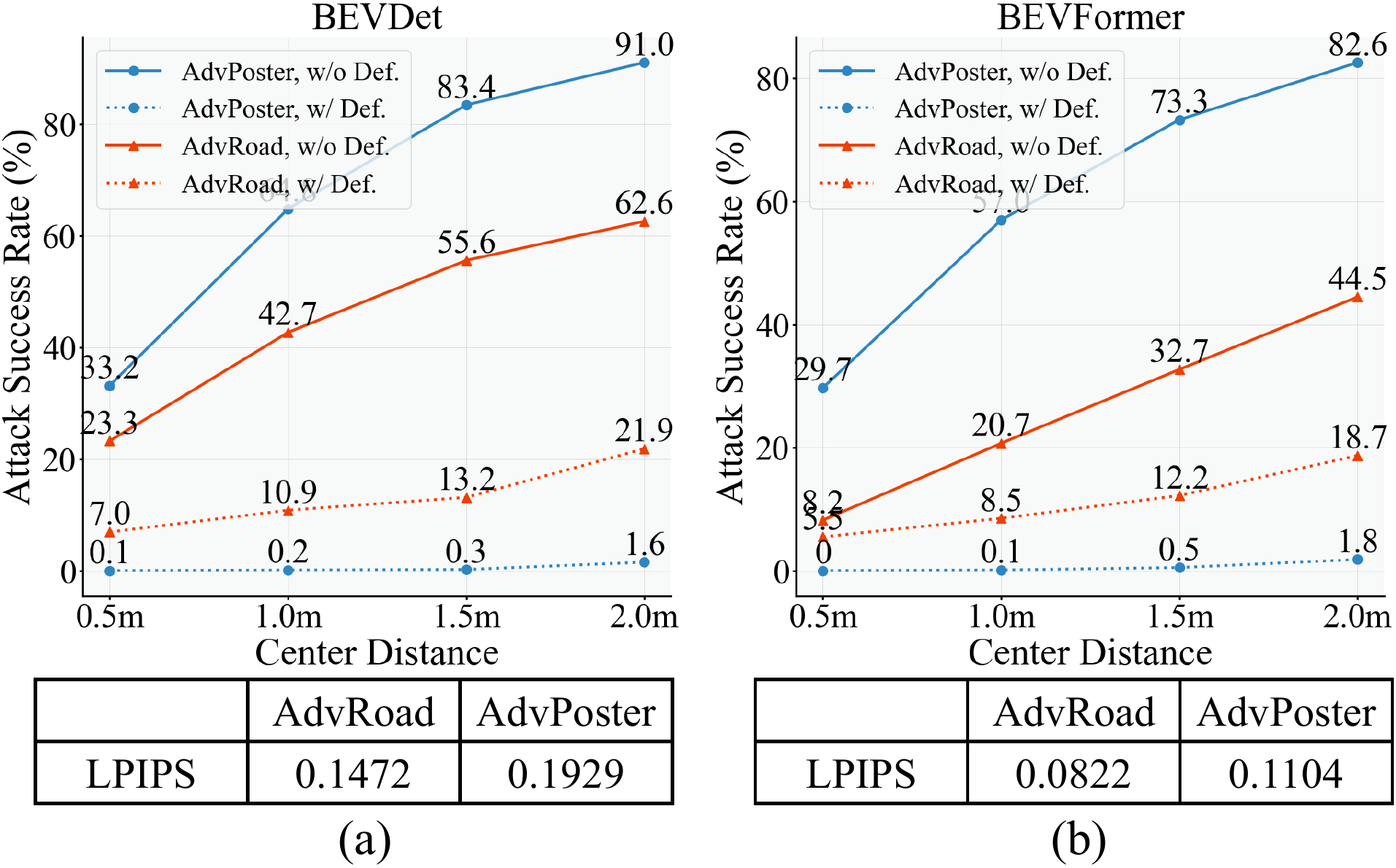}
	\caption{Comparison with AdvPoster $w/$ (dash line) and $w/o$ (solid line) defense. All victim models use ResNet50 as image backbone.}
	\label{fig_compare}
	%\Description{A woman and a girl in white dresses sit in an open car.}
\end{figure}

\textbf{Attack Performance.} The results are given in Fig. \ref{fig_compare} with aligned experimental settings (e.g., poster's physical size, attack distances). The solid lines represent the original attack performance without defense.
As shown, AdvPoster achieves superior attack performance compared to our method, attaining ASR of $91\%$ and $82.6\%$ under $CD_{2.0m}$ on BEVDet and BEVFormer respectively. This demonstrates the effectiveness of directly optimizing the explicit representation of adversaries.

\textbf{Naturality.} AdvPoster optimizes a single deceptive poster across entire input scenes. While demonstrating high attack efficacy, the learned patterns exhibit uncontrollable and often abstract characteristics. As shown in Fig. \ref{fig_main}(b), AdvPoster appears visually distinct from road surfaces and is very attention-grabbing. In contrast, our AdvRoad injects style information into the generator through implicit adversarial representations, producing natural-looking posters that blend seamlessly into road textures and achieving a lower LPIPS score. This enables stealthy attacks while amplifying practical security threats.
More quantitative visual comparisons are provided in supplementary-B.2.

\textbf{Defense Difficulty.} For AdvPoster, single training generates only one unique adversarial example with salient discriminative features, it is more likely to be targeted and defended by the AD system. Therefore, we employ adversarial augmentation as the defense. Specifically, we incorporate the learned posters into the training set and fine-tune the detector for extra 2 epochs. The attack results after defense are shown in Fig. \ref{fig_compare} (dash lines). We observe that AdvPoster only achieves less than $2\%$ ASR against the defended detectors. It is easy for the models to filter out these adversarial patterns inside the image. However, AdvRoad still achieves $\sim 20\%$ ASR under $CD_{2.0m}$, which demonstrates the strong resilience of our attack when facing the defense.  This is because AdvRoad can generate a large number of diverse adversaries, and these posters exhibit textures similar to the road surface, thus further increasing the difficulty of defense. More defense results are given in supplementary-B.3.

\subsection{Ablation}

\begin{table}[!t]
	\centering
	%\footnotesize
	\begin{tabular}{l|cc|p{0.5cm}<{\centering}p{0.5cm}<{\centering}p{0.5cm}<{\centering}p{0.5cm}<{\centering}}
		\toprule
		\multirow{2}*{}&\multicolumn{2}{c|}{\textbf{Stage}}&\multicolumn{4}{c}{\textbf{ASR ($\%, \uparrow$)}}\\ 
		&$\mathbf{1th}$&$\mathbf{2nd}$&$\mathbf{2.0m}$&$\mathbf{1.5m}$&$\mathbf{1.0m}$&$\mathbf{0.5m}$\\
		\midrule
		AdvRoad@1&$\checkmark$&&$23.4$&$19.2$&$13.1$&$6.8$\\
		AdvRoad@2&&$\checkmark$&$26.7$&$21.9$&$16.3$&$7.3$\\
		AdvRoad&$\checkmark$&$\checkmark$&$62.6$&$55.6$&$42.7$&$23.3$\\
		AdvRoad$^*$&$\checkmark$&$\checkmark$&$67.0$&$60.5$&$48.6$&$27.2$\\
		\bottomrule		
	\end{tabular}
	\caption{Ablations for Road-Style Adversary Generation and Scenario-Associated Adaptation. The results (ASR-$\%$) are given on BEVDet-R50.}
	\label{tab:ablation_two_stage}
\end{table}

\begin{table}[!t]
	\centering
	%\scriptsize
	\begin{tabular}{c|c|p{0.5cm}<{\centering}p{0.5cm}<{\centering}p{0.5cm}<{\centering}p{0.5cm}<{\centering}}
		\toprule
		\multirow{2}*{\textbf{Physical Size}}&\multirow{2}*{\textbf{LPIPS $\downarrow$}}&\multicolumn{4}{c}{\textbf{ASR ($\%, \uparrow$)}}\\ 
		&&$\mathbf{2.0m}$&$\mathbf{1.5m}$&$\mathbf{1.0m}$&$\mathbf{0.5m}$\\
		\midrule
		$1.5m\times 3.0m$&$0.1123$&$31.3$&$26.5$&$19.9$&$10.4$\\
		$2.0m\times 3.0m$&$0.1292$&$49.4$&$42.1$&$32.5$&$17.7$\\
		$2.0m\times 3.5m$&$0.1384$&$56.6$&$50.4$&$38.6$&$21.2$\\
		$2.0m\times 4.0m$&$0.1472$&$62.6$&$55.6$&$42.7$&$23.3$\\
		$2.0m\times 4.5m$&$0.1555$&$66.1$&$58.1$&$44.0$&$23.5$\\
		$2.0m\times 5.0m$&$0.1633$&$67.6$&$59.3$&$45.0$&$23.1$\\
		\bottomrule		
	\end{tabular}
	\caption{Ablations for the physical size of road posters. The results are given on BEVDet-R50.}
	\label{tab:ablation_physical_size}
\end{table}

\textbf{Two Stage Approach.}
We conduct an ablation study to validate the effectiveness of each component in the AdvRoad. The results are shown in Table \ref{tab:ablation_two_stage}. We first briefly introduce the ablation settings. AdvRoad@1 directly uses the outputs from Stage 1 for the attack; AdvRoad@2 follows the typical GAN paradigm by first training a road poster generator \emph{without the supervision of the adversarial objective $L_{obj}$}. Then, we freeze the generator using Scenario-Associated Adaptation to search the latent vector to perform the attack; AdvRoad$^*$ extends the update iteration to 50 in Stage 2. 
AdvRoad@1 achieves $23.4\%$ ASR after the Stage 1 training. This reflects the attack performance when randomly selecting a poster from the generator and placing it within the 7–10m range. 
Additionally, even with a naturally trained generator without injecting adversarial information, searching the latent space can still discover deceptive content, achieving an ASR of $26.7\%$. Combining the Road-Style Adversary Generation and Scenario-Associated Adaptation, AdvRoad boosts the ASR to $62.6\%$, which highlights the effectiveness of our two-stage attack. Also, increasing the update iterations in Stage 2 can further improve the ASR.

\textbf{Physical Size.}
Theoretically, the more pixels an attacker can manipulate in an image, the stronger attack capability becomes. However, since posters placed on road surfaces undergo perspective projection when captured by cameras, increasing the physical size of posters yields diminishing marginal returns in pixel gains for the adversarial region.
On the other hand, since AdvRoad performs an instance-level attack that aims to induce a ghost vehicle near the poster center, excessively large posters would actually hinder models from achieving precise localization. Our most lenient evaluation metric (Center distance $\leq 2m$) exactly matches the edge position of a 4m-long poster. As shown in Table \ref{tab:ablation_physical_size}, increasing the physical size can improve the ASR. However, when the poster length exceeds 4m, the incremental gains become limited. Specifically, a 5m-long poster shows a decrease in ASR under the strict evaluation criterion of $CD_{0.5m}$. However, all the posters have textures similar to the road surface, making them difficult for humans to detect.

\subsection{Attack to Broader Dataset}
\begin{figure}[!t]
	\centering
	\includegraphics[width=0.95\linewidth]{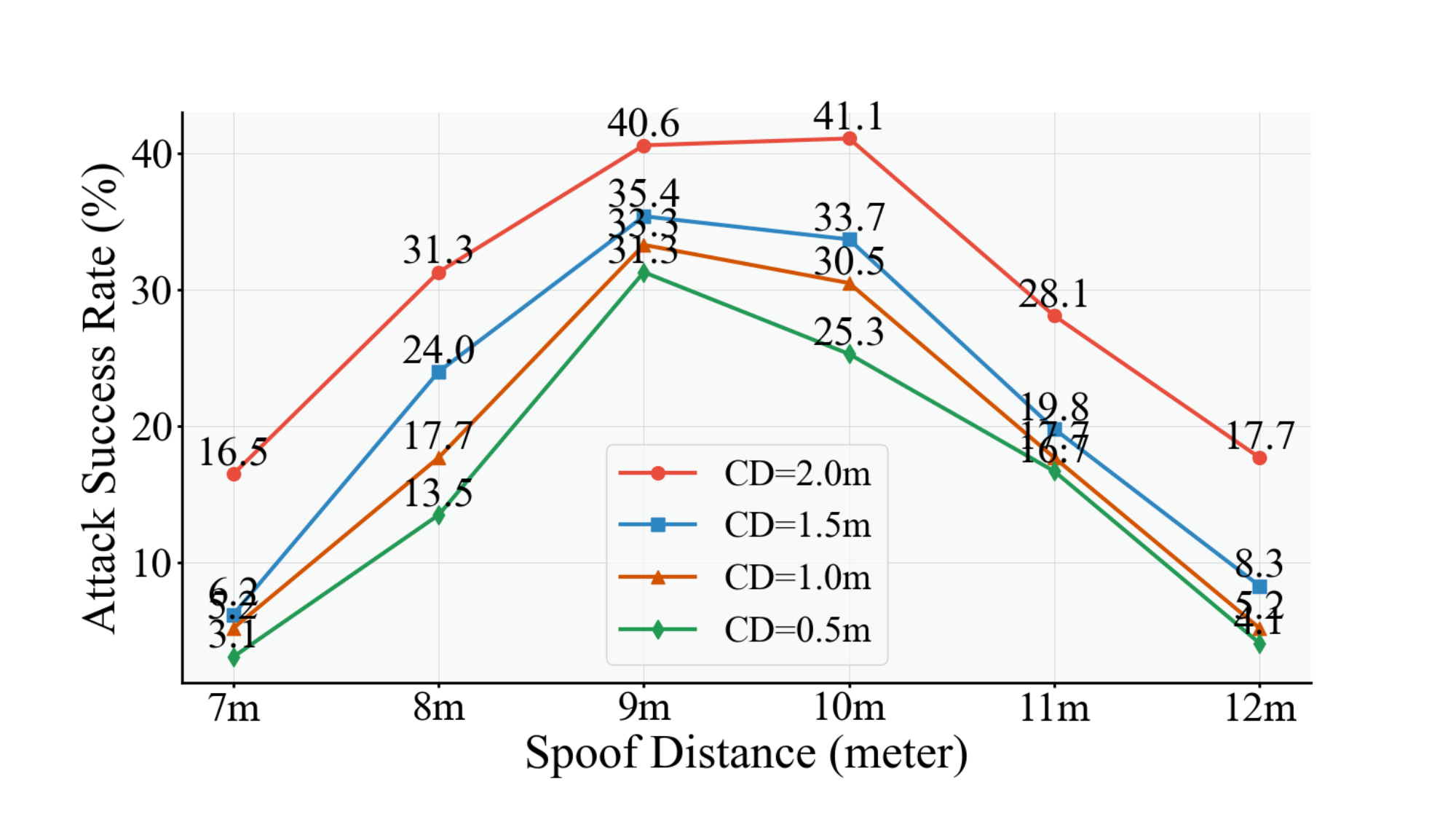}
	\caption{Attack results on the KITTI dataset. The ASRs (\%) at different spoofing distances are given.}
	\label{fig_kitti}
	%\Description{A woman and a girl in white dresses sit in an open car.}
\end{figure}

To verify the generalization ability of AdvRoad across different datasets, we further conduct attack experiments on KITTI scenes \cite{geiger2012we}. We use BEVDet as the victim model and the ASRs at different spoofing distances are shown in Fig. \ref{fig_kitti}.
We observe that the poster achieves the strongest attack effectiveness at distances between 9 and 10 meters. When the distance is too short, the poster may not be fully captured by the camera, while at longer distances, the adversarial region in the image becomes limited, leading to a decline in attack performance.

Further discussions on AdvRoad can be found in the supplementary-B.4.

\begin{figure}[!t]
	\centering
	\includegraphics[width=0.98\linewidth]{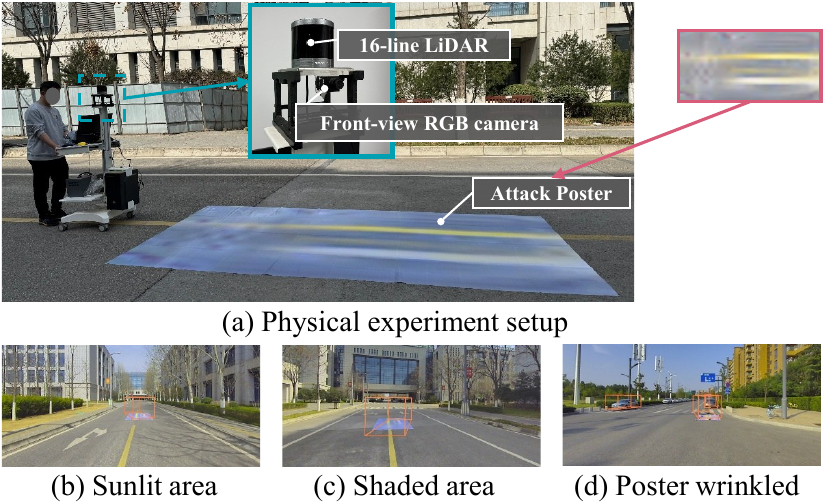}
	\caption{Physical attack environment and results. }
	\label{fig_physical_exp}
	%\Description{A woman and a girl in white dresses sit in an open car.}
\end{figure}

\subsection{Physical Attack Experiment}
To assess the practicality of our AdvRoad, we conduct attack experiments in physical-world environments. Since visual 3D detectors are typically camera-dependent (e.g., the PV2BEV transformation is related to the camera's intrinsic parameters ), physical experiments first require training a 3D detector adapted to the custom scene and camera. Therefore, we built a physical detection platform with a front-view RGB camera and a 16-line LiDAR (Fig. \ref{fig_physical_exp}(a)). The LiDAR sensor is only used to annotate the scene objects for training the custom 3D detector. Then, we train the generator, select a spoofing poster, and print it for the attack. 

Fig. \ref{fig_physical_exp}(b-d) illustrates some physical attack results. It can be seen that the poster can successfully induce the detector to generate false predictions at its location.
Despite some color deviations in the printing process (we use cost-effective banner fabric for printing), the poster remains effective. This is because, during training, we apply random brightness and contrast adjustments, as well as inject random noise, to enhance the robustness of posters. 
The quantitative results under different physical conditions are given in Table \ref{tab:phy_results}. 
In addition, the texture similar to the road surface can further reduce the attention from human drivers. 
Consider using fabrics with better color fidelity, such as canvas, to reduce the color difference with the background road surface, thereby making the attack more covert. 
Further analysis of the physical attacks is provided in the supplementary-B.5.

\begin{table}[!t]
	\centering
	\begin{tabular}{c|c}
		\toprule
		\textbf{Condition}&\textbf{ASR}\\
		\midrule
		Sunlit area&$49.4\% (170/344)$\\
		Shaded area&$28.3\% (78/276)$\\
		Poster wrinkled&$40.2\% (103/256)$\\
		Partial occlusion&$43.8\% (92/210)$\\
		Indoor&$19.5\% (57/292)$\\
		\bottomrule		
	\end{tabular}
	\caption{Quantitative attack results under different conditions.}
	\label{tab:phy_results}
\end{table}

\section{Conclusion}
This paper introduces AdvRoad, a naturalistic FP attack pipeline for visual 3D object detection in AD. AdvRoad leverages Road-Style Adversary Generation and Scenario-Associated Adaptation to perform stealthy adversarial attacks for human drivers, inducing `ghost' objects for the perception system and potentially causing real-world threats such as emergency braking. Extensive experiments on both digital and physical domains demonstrate the effectiveness of AdvRoad. Compared with previous work, our attack is harder to detect and defend against, highlighting significant security risks to AD systems.

\section{Acknowledgments}
This work was supported in part by the National Natural Science Foundation of China under Grant 62531012, in part by the National Key Research and Development Program of China under Grant 2022YFA1003800, the Key Research and Development Program of Shaanxi Province under Grant 2025CY-YBXM-040, and in part by the XJTU Research Fund for Al Science under Grant 2025YXYC004.

\bigskip

\bibliography{sample-base}

@article{ma2024vision,
  title={Vision-centric bev perception: A survey},
  author={Ma, Yuexin and Wang, Tai and Bai, Xuyang and Yang, Huitong and Hou, Yuenan and Wang, Yaming and Qiao, Yu and Yang, Ruigang and Zhu, Xinge},
  journal={IEEE Transactions on Pattern Analysis and Machine Intelligence},
  year={2024},
  publisher={IEEE}
}

@inproceedings{hu2023planning,
  title={Planning-oriented autonomous driving},
  author={Hu, Yihan and Yang, Jiazhi and Chen, Li and Li, Keyu and Sima, Chonghao and Zhu, Xizhou and Chai, Siqi and Du, Senyao and Lin, Tianwei and Wang, Wenhai and others},
  booktitle={Proceedings of the IEEE/CVF Conference on Computer Vision and Pattern Recognition},
  pages={17853--17862},
  year={2023}
}

@inproceedings{chen2017multi,
  title={Multi-view 3d object detection network for autonomous driving},
  author={Chen, Xiaozhi and Ma, Huimin and Wan, Ji and Li, Bo and Xia, Tian},
  booktitle={Proceedings of the IEEE Conference on Computer Vision and Pattern Recognition},
  pages={1907--1915},
  year={2017}
}

@article{mao20233d,
  title={3D object detection for autonomous driving: A comprehensive survey},
  author={Mao, Jiageng and Shi, Shaoshuai and Wang, Xiaogang and Li, Hongsheng},
  journal={International Journal of Computer Vision},
  volume={131},
  number={8},
  pages={1909--1963},
  year={2023},
  publisher={Springer}
}

@article{chen2024end,
  title={End-to-end autonomous driving: Challenges and frontiers},
  author={Chen, Li and Wu, Penghao and Chitta, Kashyap and Jaeger, Bernhard and Geiger, Andreas and Li, Hongyang},
  journal={IEEE Transactions on Pattern Analysis and Machine Intelligence},
  year={2024},
  publisher={IEEE}
}

@article{gulino2023waymax,
  title={Waymax: An accelerated, data-driven simulator for large-scale autonomous driving research},
  author={Gulino, Cole and Fu, Justin and Luo, Wenjie and Tucker, George and Bronstein, Eli and Lu, Yiren and Harb, Jean and Pan, Xinlei and Wang, Yan and Chen, Xiangyu and others},
  journal={Advances in Neural Information Processing Systems},
  volume={36},
  pages={7730--7742},
  year={2023}
}

@article{zhang2024visual,
  title={Visual Adversarial Attack on Vision-Language Models for Autonomous Driving},
  author={Zhang, Tianyuan and Wang, Lu and Zhang, Xinwei and Zhang, Yitong and Jia, Boyi and Liang, Siyuan and Hu, Shengshan and Fu, Qiang and Liu, Aishan and Liu, Xianglong},
  journal={arXiv preprint arXiv:2411.18275},
  year={2024}
}

@article{zhang2021evaluating,
  title={Evaluating adversarial attacks on driving safety in vision-based autonomous vehicles},
  author={Zhang, Jindi and Lou, Yang and Wang, Jianping and Wu, Kui and Lu, Kejie and Jia, Xiaohua},
  journal={IEEE Internet of Things Journal},
  volume={9},
  number={5},
  pages={3443--3456},
  year={2021},
  publisher={IEEE}
}

@inproceedings{zhu2023understanding,
  title={Understanding the robustness of 3D object detection with bird's-eye-view representations in autonomous driving},
  author={Zhu, Zijian and Zhang, Yichi and Chen, Hai and Dong, Yinpeng and Zhao, Shu and Ding, Wenbo and Zhong, Jiachen and Zheng, Shibao},
  booktitle={Proceedings of the IEEE/CVF Conference on Computer Vision and Pattern Recognition},
  pages={21600--21610},
  year={2023}
}

@inproceedings{abdelfattah2021adversarial,
  title={Adversarial attacks on camera-lidar models for 3d car detection},
  author={Abdelfattah, Mazen and Yuan, Kaiwen and Wang, Z Jane and Ward, Rabab},
  booktitle={IEEE/RSJ International Conference on Intelligent Robots and Systems},
  pages={2189--2194},
  year={2021},
  organization={IEEE}
}

@article{cheng2023fusion,
  title={Fusion is not enough: Single modal attacks on fusion models for 3D object detection},
  author={Cheng, Zhiyuan and Choi, Hongjun and Liang, James and Feng, Shiwei and Tao, Guanhong and Liu, Dongfang and Zuzak, Michael and Zhang, Xiangyu},
  journal={arXiv preprint arXiv:2304.14614},
  year={2023}
}

@article{zhang2024comprehensive,
  title={A comprehensive study of the robustness for lidar-based 3d object detectors against adversarial attacks},
  author={Zhang, Yifan and Hou, Junhui and Yuan, Yixuan},
  journal={International Journal of Computer Vision},
  volume={132},
  number={5},
  pages={1592--1624},
  year={2024},
  publisher={Springer}
}

@article{wang2025unified,
  title={A Unified Framework for Adversarial Patch Attacks against Visual 3D Object Detection in Autonomous Driving},
  author={Wang, Jian and Li, Fan and He, Lijun},
  journal={IEEE Transactions on Circuits and Systems for Video Technology},
  year={2025},
  publisher={IEEE}
}

@article{wang2025physically,
  title={Physically Realizable Adversarial Creating Attack against Vision-based BEV Space 3D Object Detection},
  author={Wang, Jian and Li, Fan and Lv, Song and He, Lijun and Shen, Chao},
  journal={IEEE Transactions on Image Processing},
  year={2025},
  publisher={IEEE}
}

@inproceedings{cao2023you,
  title={You can't see me: Physical removal attacks on $\{$lidar-based$\}$ autonomous vehicles driving frameworks},
  author={Cao, Yulong and Bhupathiraju, S Hrushikesh and Naghavi, Pirouz and Sugawara, Takeshi and Mao, Z Morley and Rampazzi, Sara},
  booktitle={USENIX Security Symposium},
  pages={2993--3010},
  year={2023}
}

@inproceedings{jin2023pla,
  title={Pla-lidar: Physical laser attacks against lidar-based 3d object detection in autonomous vehicle},
  author={Jin, Zizhi and Ji, Xiaoyu and Cheng, Yushi and Yang, Bo and Yan, Chen and Xu, Wenyuan},
  booktitle={IEEE Symposium on Security and Privacy},
  pages={1822--1839},
  year={2023},
  organization={IEEE}
}

@article{hau2021object,
  title={Object removal attacks on lidar-based 3d object detectors},
  author={Hau, Zhongyuan and Co, Kenneth T and Demetriou, Soteris and Lupu, Emil C},
  journal={arXiv preprint arXiv:2102.03722},
  year={2021}
}

@inproceedings{sun2020towards,
  title={Towards robust LiDAR-based perception in autonomous driving: General black-box adversarial sensor attack and countermeasures},
  author={Sun, Jiachen and Cao, Yulong and Chen, Qi Alfred and Mao, Z Morley},
  booktitle={USENIX Security Symposium},
  pages={877--894},
  year={2020}
}

@inproceedings{cao2019adversarial,
  title={Adversarial sensor attack on lidar-based perception in autonomous driving},
  author={Cao, Yulong and Xiao, Chaowei and Cyr, Benjamin and Zhou, Yimeng and Park, Won and Rampazzi, Sara and Chen, Qi Alfred and Fu, Kevin and Mao, Z Morley},
  booktitle={Proceedings of the ACM SIGSAC Conference on Computer and Communications Security},
  pages={2267--2281},
  year={2019}
}

@article{wang2023adversarial,
  title={Adversarial obstacle generation against lidar-based 3d object detection},
  author={Wang, Jian and Li, Fan and Zhang, Xuchong and Sun, Hongbin},
  journal={IEEE Transactions on Multimedia},
  volume={26},
  pages={2686--2699},
  year={2023},
  publisher={IEEE}
}

@inproceedings{tu2020physically,
  title={Physically realizable adversarial examples for lidar object detection},
  author={Tu, James and Ren, Mengye and Manivasagam, Sivabalan and Liang, Ming and Yang, Bin and Du, Richard and Cheng, Frank and Urtasun, Raquel},
  booktitle={Proceedings of the IEEE/CVF Conference on Computer Vision and Pattern Recognition},
  pages={13716--13725},
  year={2020}
}

@article{cao2019adversarialv2,
  title={Adversarial objects against lidar-based autonomous driving systems},
  author={Cao, Yulong and Xiao, Chaowei and Yang, Dawei and Fang, Jing and Yang, Ruigang and Liu, Mingyan and Li, Bo},
  journal={arXiv preprint arXiv:1907.05418},
  year={2019}
}

@INPROCEEDINGS{li2023adv3d,
  author={Li, Leheng and Lian, Qing and Chen, Ying-Cong},
  booktitle={IEEE/RSJ International Conference on Intelligent Robots and Systems}, 
  title={Adv3D: Generating 3D Adversarial Examples for 3D Object Detection in Driving Scenarios with NeRF}, 
  year={2024},
  volume={},
  number={},
  pages={10813-10820},
  doi={10.1109/IROS58592.2024.10801323}}

@article{goodfellow2014explaining,
  title={Explaining and harnessing adversarial examples},
  author={Goodfellow, Ian J and Shlens, Jonathon and Szegedy, Christian},
  journal={arXiv preprint arXiv:1412.6572},
  year={2014}
}

@inproceedings{athalye2018synthesizing,
  title={Synthesizing robust adversarial examples},
  author={Athalye, Anish and Engstrom, Logan and Ilyas, Andrew and Kwok, Kevin},
  booktitle={International Conference on Machine Learning},
  pages={284--293},
  year={2018},
  organization={PMLR}
}

@article{yang2025adversarial,
  title={Adversarial Example Soups: Improving Transferability and Stealthiness for Free},
  author={Yang, Bo and Zhang, Hengwei and Wang, Jindong and Yang, Yulong and Lin, Chenhao and Shen, Chao and Zhao, Zhengyu},
  journal={IEEE Transactions on Information Forensics and Security},
  year={2025},
  publisher={IEEE}
}

@article{lin2024hard,
  title={Hard Adversarial Example Mining for Improving Robust Fairness},
  author={Lin, Chenhao and Ji, Xiang and Yang, Yulong and Li, Qian and Zhao, Zhengyu and Peng, Zhe and Wang, Run and Fang, Liming and Shen, Chao},
  journal={IEEE Transactions on Information Forensics and Security},
  year={2024},
  publisher={IEEE}
}

@article{han2023interpreting,
  title={Interpreting adversarial examples in deep learning: A review},
  author={Han, Sicong and Lin, Chenhao and Shen, Chao and Wang, Qian and Guan, Xiaohong},
  journal={ACM Computing Surveys},
  volume={55},
  number={14s},
  pages={1--38},
  year={2023},
  publisher={ACM New York, NY}
}

@inproceedings{wang2023towards,
  title={Towards transferable targeted adversarial examples},
  author={Wang, Zhibo and Yang, Hongshan and Feng, Yunhe and Sun, Peng and Guo, Hengchang and Zhang, Zhifei and Ren, Kui},
  booktitle={Proceedings of the IEEE/CVF Conference on Computer Vision and Pattern Recognition},
  pages={20534--20543},
  year={2023}
}

@inproceedings{guesmi2024dap,
  title={Dap: A dynamic adversarial patch for evading person detectors},
  author={Guesmi, Amira and Ding, Ruitian and Hanif, Muhammad Abdullah and Alouani, Ihsen and Shafique, Muhammad},
  booktitle={Proceedings of the IEEE/CVF Conference on Computer Vision and Pattern Recognition},
  pages={24595--24604},
  year={2024}
}

@inproceedings{thys2019fooling,
  title={Fooling automated surveillance cameras: adversarial patches to attack person detection},
  author={Thys, Simen and Van Ranst, Wiebe and Goedem{\'e}, Toon},
  booktitle={Proceedings of the IEEE/CVF Conference on Computer Vision and Pattern Recognition Workshops},
  pages={0--0},
  year={2019}
}

@article{hu2024adversarial,
  title={Adversarial infrared curves: An attack on infrared pedestrian detectors in the physical world},
  author={Hu, Chengyin and Shi, Weiwen and Yao, Wen and Jiang, Tingsong and Tian, Ling and Chen, Xiaoqian and Li, Wen},
  journal={Neural Networks},
  volume={178},
  pages={106459},
  year={2024},
  publisher={Elsevier}
}

@article{hu2025two,
  title={Two-stage optimized unified adversarial patch for attacking visible-infrared cross-modal detectors in the physical world},
  author={Hu, Chengyin and Shi, Weiwen and Yao, Wen and Jiang, Tingsong and Tian, Ling and Li, Wen},
  journal={Applied Soft Computing},
  pages={112818},
  year={2025},
  publisher={Elsevier}
}

@article{lee2019physical,
  title={On physical adversarial patches for object detection},
  author={Lee, Mark and Kolter, Zico},
  journal={arXiv preprint arXiv:1906.11897},
  year={2019}
}

@inproceedings{hu2021naturalistic,
  title={Naturalistic physical adversarial patch for object detectors},
  author={Hu, Yu-Chih-Tuan and Kung, Bo-Han and Tan, Daniel Stanley and Chen, Jun-Cheng and Hua, Kai-Lung and Cheng, Wen-Huang},
  booktitle={Proceedings of the IEEE/CVF International Conference on Computer Vision},
  pages={7848--7857},
  year={2021}
}

@article{xie2023adversarial,
  title={On the adversarial robustness of camera-based 3d object detection},
  author={Xie, Shaoyuan and Li, Zichao and Wang, Zeyu and Xie, Cihang},
  journal={arXiv preprint arXiv:2301.10766},
  year={2023}
}

@article{huang2021bevdet,
  title={Bevdet: High-performance multi-camera 3d object detection in bird-eye-view},
  author={Huang, Junjie and Huang, Guan and Zhu, Zheng and Ye, Yun and Du, Dalong},
  journal={arXiv preprint arXiv:2112.11790},
  year={2021}
}

@article{huang2022bevdet4d,
  title={Bevdet4d: Exploit temporal cues in multi-camera 3d object detection},
  author={Huang, Junjie and Huang, Guan},
  journal={arXiv preprint arXiv:2203.17054},
  year={2022}
}

@article{li2024bevformer,
  title={Bevformer: learning bird's-eye-view representation from lidar-camera via spatiotemporal transformers},
  author={Li, Zhiqi and Wang, Wenhai and Li, Hongyang and Xie, Enze and Sima, Chonghao and Lu, Tong and Yu, Qiao and Dai, Jifeng},
  journal={IEEE Transactions on Pattern Analysis and Machine Intelligence},
  year={2024},
  publisher={IEEE}
}

@inproceedings{he2016deep,
  title={Deep residual learning for image recognition},
  author={He, Kaiming and Zhang, Xiangyu and Ren, Shaoqing and Sun, Jian},
  booktitle={Proceedings of the IEEE Conference on Computer Vision and Pattern Recognition},
  pages={770--778},
  year={2016}
}

@inproceedings{liu2021swin,
  title={Swin transformer: Hierarchical vision transformer using shifted windows},
  author={Liu, Ze and Lin, Yutong and Cao, Yue and Hu, Han and Wei, Yixuan and Zhang, Zheng and Lin, Stephen and Guo, Baining},
  booktitle={Proceedings of the IEEE/CVF International Conference on Computer Vision},
  pages={10012--10022},
  year={2021}
}

@inproceedings{caesar2020nuscenes,
  title={nuscenes: A multimodal dataset for autonomous driving},
  author={Caesar, Holger and Bankiti, Varun and Lang, Alex H and Vora, Sourabh and Liong, Venice Erin and Xu, Qiang and Krishnan, Anush and Pan, Yu and Baldan, Giancarlo and Beijbom, Oscar},
  booktitle={Proceedings of the IEEE/CVF Conference on Computer Vision and Pattern Recognition},
  pages={11621--11631},
  year={2020}
}

@inproceedings{geiger2012we,
  title={Are we ready for autonomous driving? the kitti vision benchmark suite},
  author={Geiger, Andreas and Lenz, Philip and Urtasun, Raquel},
  booktitle={IEEE Conference on Computer Vision and Pattern Recognition},
  pages={3354--3361},
  year={2012},
  organization={IEEE}
}

@inproceedings{zhang2018unreasonable,
  title={The unreasonable effectiveness of deep features as a perceptual metric},
  author={Zhang, Richard and Isola, Phillip and Efros, Alexei A and Shechtman, Eli and Wang, Oliver},
  booktitle={Proceedings of the IEEE Conference on Computer Vision and Pattern Recognition},
  pages={586--595},
  year={2018}
}

@inproceedings{sato2021dirty,
  title={Dirty road can attack: Security of deep learning based automated lane centering under Physical-World attack},
  author={Sato, Takami and Shen, Junjie and Wang, Ningfei and Jia, Yunhan and Lin, Xue and Chen, Qi Alfred},
  booktitle={USENIX security symposium},
  pages={3309--3326},
  year={2021}
}

\end{document}

\maketitle

\section{A. Details of Image-3D Rendering}
In this section, we present the details of the Image-3D rendering algorithm that differentiably render the 3D space poster onto the image. In general, the core of the algorithm lies in finding the relative position of each pixel in the adversarial image region on the predefined poster, enabling per-pixel interpolation.

Specifically, we first calculate the projection matrix $M_{3D\rightarrow img} \in \mathbb{R}^{4\times 4}$ according to the camera's intrinsic and extrinsic parameters. $M_{3D\rightarrow img}$ can project points from the 3D LiDAR coordinate system to the image plane.
Then, given four poster corners $\{p^{3d}_i \in \mathbb{R}^3|i=1,2,3,4\}$ in 3D coordinate system, we get four projected corner points $\{p^{img}_i\in \mathbb{R}^2|i=1,2,3,4\}$ in image plane:
\begin{equation}
	\begin{aligned}
		[u_i,v_i,d_i,1]^T &= M_{3D\rightarrow img}[x_i,y_i,z_i,1]^T\\
		p^{img}_i &= (u_i/d_i, v_i/d_i)
	\end{aligned}
\end{equation}
The adversarial image region is the intersection between the quadrangle region defined by the projected corner points and the image. For each pixel in this region, we need to compute its position in the 3D space. However, we cannot compute it directly due to the lack of per-pixel depth information $d_i$. Nevertheless, we can indirectly calculate the 3D position of each point by leveraging the known height information $z_{height}$ of the poster. Formally, for an adversarial point $p^{img}_i = (u_i,v_i)$, the corresponding 3D point $p^{3d}_i=(x_i,y_i,z_i)$ can be solved:
\begin{equation}
	\begin{aligned}
		[x_i,y_i,z_i,1]^T &= M_{3D\rightarrow img}^{-1}[u_id_i,v_id_i,d_i,1]^T\\
		z_i&=z_{height}
	\end{aligned}
\end{equation}
Finally, the color pixel of $p^{img}_i$ is interpolated across the neighbor values around $p^{3d}_i$ on the poster.

\begin{figure}[!t]
	\centering
	\includegraphics[width=\linewidth]{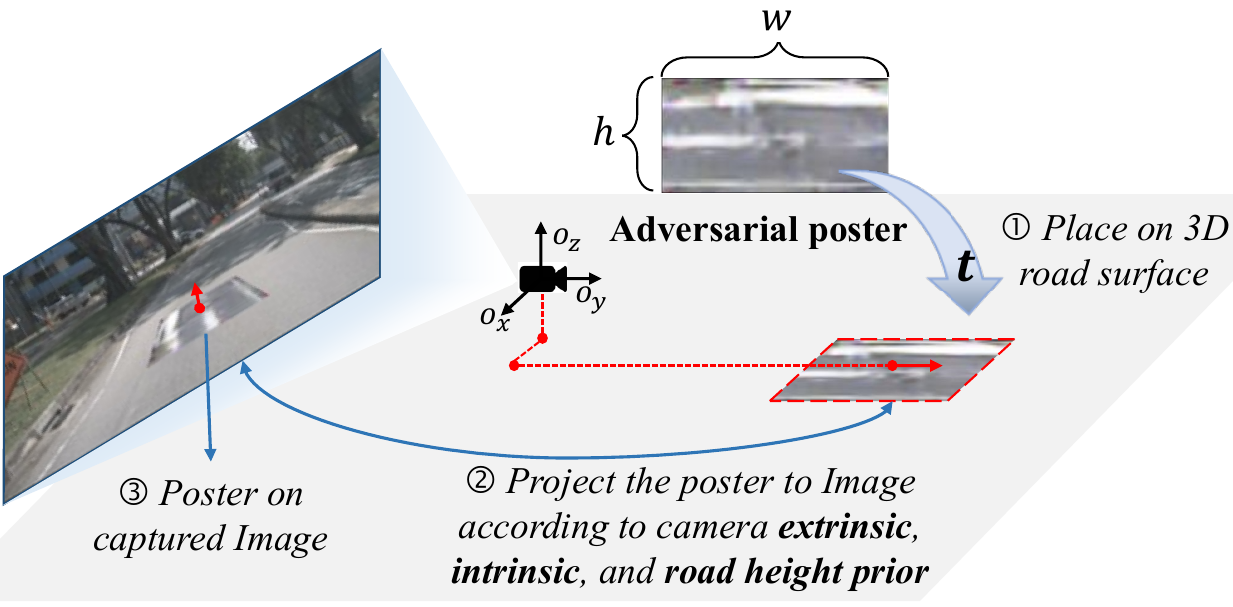}
	\caption{Illustration of Image-3D rendering.}
	\label{fig_render}
	%\Description{A woman and a girl in white dresses sit in an open car.}
\end{figure}

\begin{table}[h]
	\footnotesize
	\begin{tabular}{c|cc|cc}
		\hline
		\textbf{Model}&\textbf{Backbone}&\textbf{Input Resolution}&\textbf{mAP}&\textbf{NDS}\\
		\hline
		\multirow{2}*{BEVDet}&ResNet50&$256\times 704$&$31.7$&$39.4$\\
		&Swin-Tiny&$256\times 704$&$33.4$&$41.8$\\
		\hline
		BEVDet4D&ResNet50&$256\times 704$&$34.6$&$48.0$\\
		(8 Frames)&Swin-Tiny&$256\times 704$&$36.3$&$48.9$\\
		\hline
		BEVFormer&ResNet50&$480\times 800$&$26.7$&$36.9$\\
		-Tiny&Swin-Tiny&$480\times 800$&$27.1$&$37.8$\\
		\hline
	\end{tabular}
	\caption{Detection performance of all victim detectors on the nuScenes validation set.\label{detector_performance}}
	\centering
\end{table}

\begin{figure*}[!h]
	\centering
	\includegraphics[width=0.75\linewidth]{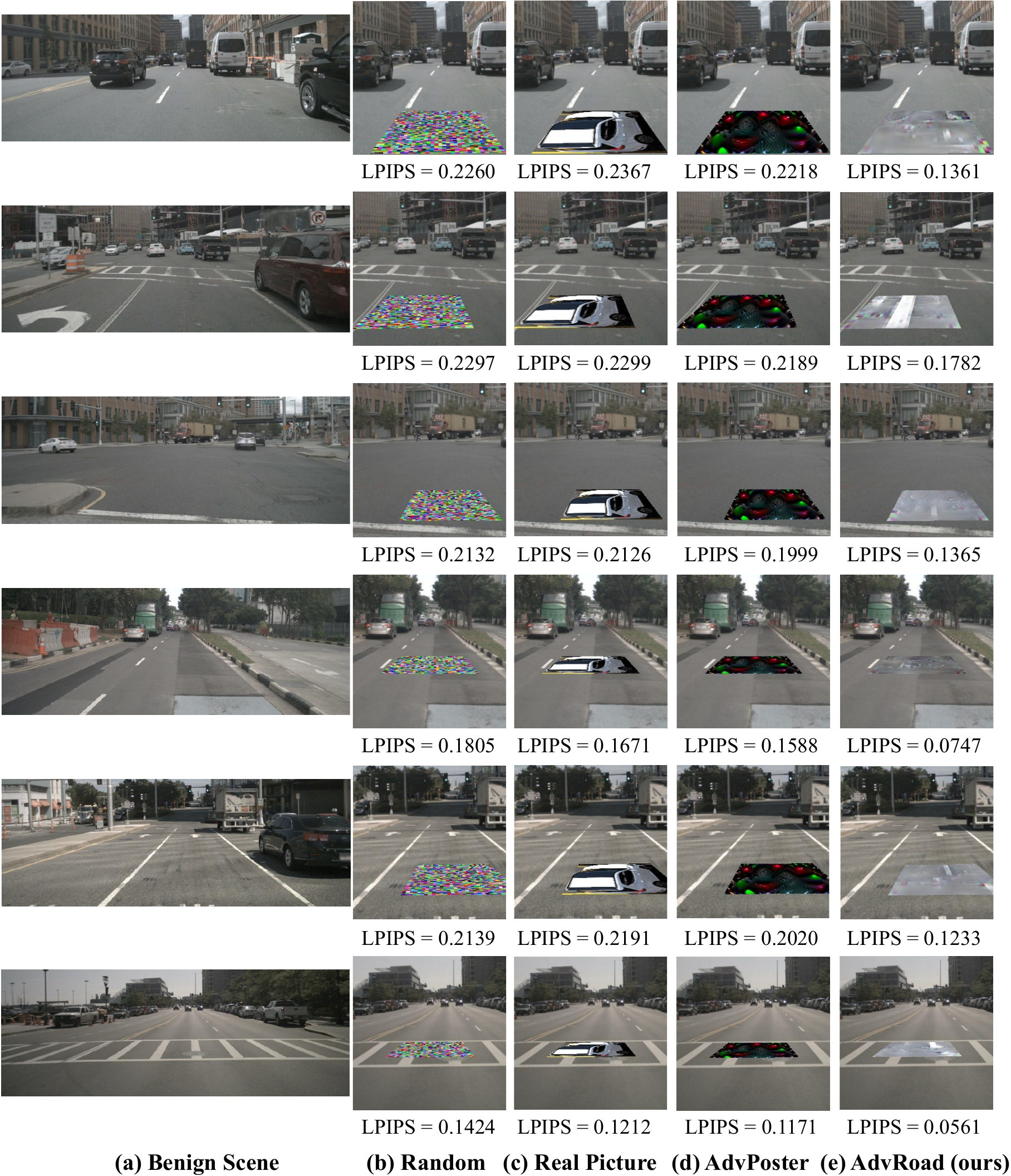}
	\caption{Quantitative visual comparison in terms of the LPIPS ($\downarrow$) with other attack posters.}
	\label{fig_lpips}
	%\Description{A woman and a girl in white dresses sit in an open car.}
\end{figure*}

\section{B. Additional Experimental Results}
\noindent\textbf{Training Detail.} In stage 1, we train the generator and discriminator alternately with an iteration ratio $1:10$, using Adam optimizer with a learning rate of 0.0001. The generator is totally updated for 16 epochs on the nuScenes training set with a batchsize of 40 (we accumulate the gradients of mini-batch multiple times to achieve a larger batchsize to stabilize training).  The weight factor $\lambda = 0.1$. 
Stage 2 updates the randomly initialized latent vector 30 iterations for each input frame.

\subsection{B.1 Detection Performance of Victim Models}

We present the detection performance of all six victim 3D detectors in Table \ref{detector_performance}. For the BEVDet and BEVDet4D, we train the model on $4$ NVIDIA GeForce RTX 3090 GPUs with learning rate 2e-4 for $20$ epochs (with CBGS) and drop the learning rate at epoch $17$ by a factor of $0.1$. The BEV grid size is $0.6m\times0.6m$ and $0.8m\times0.8m$ for BEVDet and BEVDet4D respectively. For the BEVFormer, we train the model for $24$ epochs with the CosineAnnealing strategy (initial learning rate is 2e-4). The BEV grid size is $0.512m\times 0.512m$.

\subsection{B.2 Visualization of Different Attack Posters}
We provide more visual comparisons with other attack posters in Fig. \ref{fig_lpips}. The LPIPS score ($\downarrow$) evaluates the perceptual similarity by utilizing neural networks to compare multi-layer features between benign and attacked images.  The results show that AdvRoad exhibits the best concealment and environmental consistency. 

\subsection{B.3 Adversarial Segmentation as Defense}
\begin{table}[!t]
	\centering
	\begin{tabular}{c|cccc}
		\toprule
		\multirow{2}*{\textbf{Attack}}&\multicolumn{4}{c}{\textbf{ASR ($\%, \uparrow$)}}\\ 
		&$\mathbf{2.0m}$&$\mathbf{1.5m}$&$\mathbf{1.0m}$&$\mathbf{0.5m}$\\
		\midrule
		AdvPoster&$19.7\%$&$14.6\%$&$10.4\%$&$3.9\%$\\
		AdvRoad&$32.4\%$&$25.5\%$&$14.8\%$&$7.1\%$\\
		\bottomrule		
	\end{tabular}
	\caption{Attack results after applying adversarial segmentation as defense.}
	\label{tab:adv_seg}
\end{table}

We train a lightweight adversarial segmentation network to predict the poster areas in the image. During evaluation, the poster region in the input image is first segmented and masked out before performing detection. The resulting ASR on BEVDet-R50 after applying this defense is shown in Table \ref{tab:adv_seg}. 
Although AdvRoad has a lower ASR than AdvPoster without defense, it shows stronger resistance to the fine-tuned detector and adversarial segmentation. This robustness comes from: 1) its similarity to the road background, and 2) diverse poster content.

\subsection{B.4 Discussion}
\begin{table*}
	\centering
	\begin{tabular}{c|c|cc|c|cccc}
		\toprule
		\multirow{2}*{\textbf{Model}}&\multirow{2}*{\textbf{Attack}}&\multirow{2}*{\textbf{Diversity}}&\multirow{2}*{\textbf{Natural-looking}}&\multirow{2}*{\textbf{LPIPS $\downarrow$}}&\multicolumn{4}{c}{\textbf{ASR ($\%, \uparrow$)}}\\ 
		&&&&&$\mathbf{2.0m}$&$\mathbf{1.5m}$&$\mathbf{1.0m}$&$\mathbf{0.5m}$\\
		\midrule
		\multirow{3}*{BEVDet-R50}&AdvPoster&$\times$&$\times$&$0.1929$&$\mathbf{91.0}$&$\mathbf{83.4}$&$\mathbf{64.8}$&$\mathbf{33.2}$\\
		&AdvRoad&$\checkmark$&$\checkmark$& $\mathbf{0.1472}$& $62.6$& $55.6$& $42.7$& $23.3$\\
		&AdvRoad$\dagger$&$\times$&$\times$&$0.2100$&$80.3$&$69.4$&$49.4$&$22.1$\\
		\midrule
		\multirow{3}*{BEVDet-SwinT}&AdvPoster&$\times$&$\times$&$0.1737$&$\mathbf{91.8}$&$\mathbf{84.7}$&$66.5$&$38.3$\\
		&AdvRoad&$\checkmark$&$\checkmark$& $\mathbf{0.1337}$& $60.2$&$56.3$&$47.6$&$28.8$\\
		&AdvRoad$\dagger$&$\times$&$\times$&$0.2172$&$86.7$&$81.5$&$\mathbf{69.2}$&$\mathbf{41.9}$\\
		\midrule
		\multirow{3}*{BEVDet4D-R50}&AdvPoster&$\times$&$\times$&$0.1758$&$\mathbf{86.9}$&$\mathbf{78.6}$&$\mathbf{58.7}$&$\mathbf{30.8}$\\
		&AdvRoad&$\checkmark$&$\checkmark$&$\mathbf{0.1331}$&$49.1$&$42.9$& $32.7$&$17.7$\\
		&AdvRoad$\dagger$&$\times$&$\times$&$0.2223$&$79.0$&$71.8$&$55.6$&$29.6$\\
		\midrule
		\multirow{3}*{BEVDet4D-SwinT}&AdvPoster&$\times$&$\times$&$0.1748$&$\mathbf{92.7}$&$\mathbf{87.9}$&$\mathbf{73.4}$&$\mathbf{40.8}$\\
		&AdvRoad&$\checkmark$&$\checkmark$&$\mathbf{0.1370}$&$39.1$&$35.1$&$27.8$&$15.7$\\
		&AdvRoad$\dagger$&$\times$&$\times$&$0.1966$&$86.8$&$80.5$&$64.9$&$36.6$\\
		\midrule
		\multirow{3}*{BEVFormer-R50}&AdvPoster&$\times$&$\times$&$0.1104$&$\mathbf{82.6}$&$\mathbf{73.3}$&$\mathbf{57.0}$&$\mathbf{29.7}$\\
		&AdvRoad&$\checkmark$&$\checkmark$&$\mathbf{0.0822}$&$44.5$&$32.7$&$20.7$&$8.2$\\
		&AdvRoad$\dagger$&$\times$&$\times$&$0.1082$&$70.1$&$58.0$&$41.2$&$19.8$\\
		\midrule
		\multirow{3}*{BEVFormer-SwinT}&AdvPoster&$\times$&$\times$&$0.1026$&$\mathbf{83.9}$&$\mathbf{74.8}$&$\mathbf{58.0}$&$\mathbf{31.0}$\\
		&AdvRoad&$\checkmark$&$\checkmark$&$\mathbf{0.0818}$&$37.3$&$30.6$&$21.0$&$8.9$\\
		&AdvRoad$\dagger$&$\times$&$\times$&$0.1196$&$74.9$&$65.4$&$48.5$&$24.3$\\
		\bottomrule
		%&Side 
		%&\multicolumn{4}{c}{\textbf{Scene Creating Attack}}

	\end{tabular}
	\caption{Digital attack results of different creation posters in the nuScenes dataset. $\dagger$ denotes training the generator under the supervision of adversarial objective $L_{obj}$ only.}
	\label{tab:exp_adv_obj}
\end{table*}

\textbf{Transferability.}
We evaluate the transferability of AdvRoad across different detectors through the black-box settings, where attackers have no access to the details of the target model. Since the latent vector search in Stage 2 requires interaction between the adversary input and the model, we only use posters generated in Stage 1 for the attack. Specifically, we first use the generator trained on the source model to randomly generate 500 posters, then select the 10 most effective ones (with the highest confidence) to attack the target models. This process does not require any information about the target models. 

\begin{figure}[!h]
	\centering
	\includegraphics[width=\linewidth]{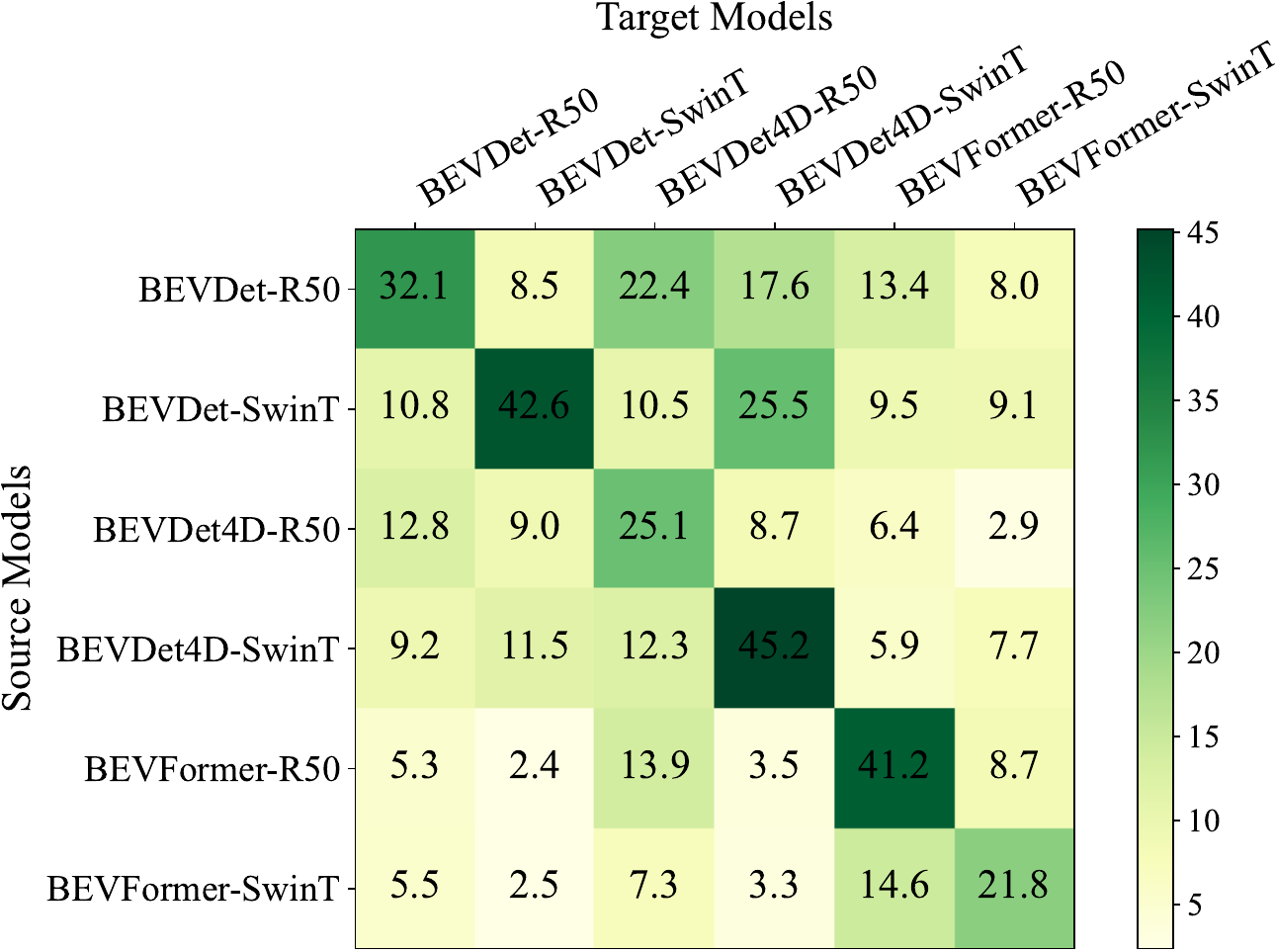}
	\caption{Transfer results across six detectors. The ASR ($\%$) under $CD_{2.0m}$ are given.}
	\label{fig_transfer}
	%\Description{A woman and a girl in white dresses sit in an open car.}
\end{figure}

The transfer results are provided in Fig. \ref{fig_transfer}. We make the following observations. First, detectors that adopt the same PV2BEV transformation exhibit better transferability. For example, the four models within the BEVDet series achieve a minimum ASR of $8.5\%$, while the posters generated on BEVFormer showed a lower limit on BEVDet. Such as a minimum of $2.4\%$, which almost fails in this case. Since the BEVDet series requires explicit per-pixel depth estimation for feature projection, whereas BEVFormer does not, this may lead to differences in the semantic focus of the learned posters. Second, detectors with the same backbone exhibit better transferability. For example, posters learned on BEVDet-SwinT achieve $25.5\%$ ASR on BEVDet4D-SwinT, compared with $10.5\%$ on BEVDet4D-R50.
These results may provide insights for designing more transferable adversarial attacks.

\begin{figure}[!t]
	\centering
	\includegraphics[width=\linewidth]{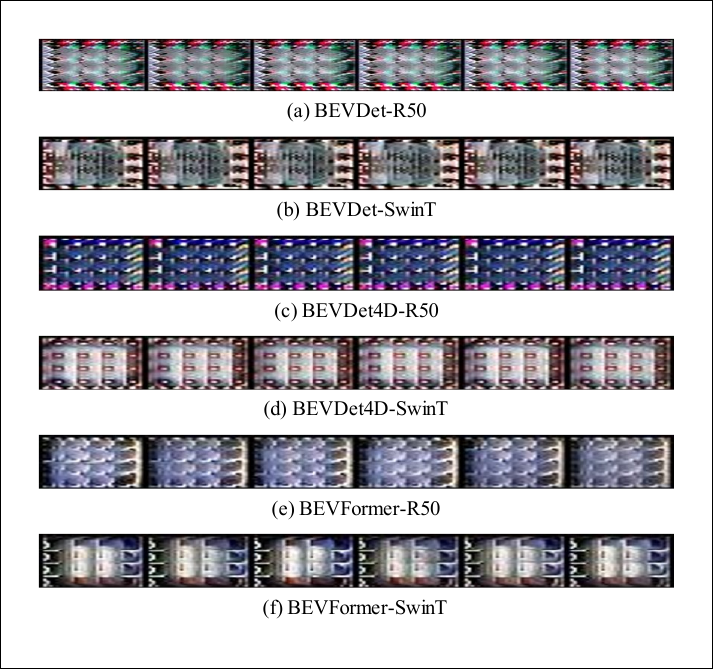}
	\caption{Visualization of the posters for AdvRoad$\dagger$. The posters from the same model (randomly generating 6 samples for each model) share similar content and have unnatural patterns.}
	\label{fig_vis_gan_adv_obj}
	%\Description{A woman and a girl in white dresses sit in an open car.}
\end{figure}

\textbf{Diversity and Effectiveness.} To explore the upper bound of the attack capability of the implicit representation of the adversarial poster, we supervise the generator using only $L_{obj}$, without considering gradients from $L_{cls}$. The results are given in Table \ref{tab:exp_adv_obj}. It can be observed that AdvRoad$\dagger$ achieves a significant improvement in ASR compared to the AdvRoad, at the cost of natural appearance.  Some of the poster output from the generator is visualized in Fig. \ref{fig_vis_gan_adv_obj}. We find that no matter what the latent vector input is, the generator outputs similar content and loses the diversity. Moreover, the explicit representation of the adversary, AdvPoster, demonstrates stronger attack capability and higher optimization efficiency. Similarly, at the cost of diversity and visual naturalness.

\subsection{B.5 Further Analysis of the Physical Attack}

In practice, our physical experiments yield the following insights:
1. Factors affecting overall color—such as lighting, print color deviation, and reflectivity—are key to attack success. Due to AdvRoad’s similarity to road surfaces, its deceptive patterns require high color fidelity. In contrast, AdvPoster’s vivid patterns show greater robustness to such factors.
2. Local distortions (e.g., wrinkles, occlusions) usually cause only positional shifts in detection results and do not invalidate the attack. This suggests that the adversarial information is distributed across the whole poster and is insensitive to local damage.
3. Although we didn’t test under rain or night conditions, we suspect both methods would perform poorly, as much information would be lost after image capture. For example, 3D detectors already struggle to detect real vehicles at night. A potential solution is to use projectors to display the poster, which we plan to explore in future work.

%\bibliography{sample-base}